\documentclass[letterpaper]{article} 
\usepackage[preprint]{aaai2027}
\usepackage[hyphens]{url}  
\usepackage{graphicx} 
\usepackage{natbib}  
\usepackage{caption} 
\usepackage{algorithm}
\usepackage{algorithmic}
\usepackage{multirow}
\usepackage{amsfonts}
\usepackage{pifont}

\usepackage{newfloat}
\usepackage{listings}
\DeclareCaptionStyle{ruled}{labelfont=normalfont,labelsep=colon,strut=off} 
\floatstyle{ruled}
\newfloat{listing}{tb}{lst}{}
\floatname{listing}{Listing}

\usepackage{booktabs}

\title{Distilling Vision-Language Models for On-Device Fire Understanding}
\author{
    Mohammad Kazzazi \textsuperscript{\rm 1}\equalcontrib,
    Zixuan Liu\textsuperscript{\rm 2}\equalcontrib\corresponding,
    Siavash Khajavi\textsuperscript{\rm 2}\corresponding
}
\affiliations{
    \textsuperscript{\rm 1}EPFL, Lausanne, Switzerland \\
    \textsuperscript{\rm 2}Detectium, Espoo, Finland\\

    seyedmohammad.ojaghkazazi@epfl.ch, zixuan@detectium.io, siavash.khajavi@aalto.fi
}

\begin{document}

\maketitle

\begin{abstract}
Vision-language models (VLMs) offer a promising alternative to conventional fire detection systems by reasoning about the semantic context of a scene and thus reducing false alarms, yet their large model size makes deployment on embedded fire sensors impractical. In this paper, we study how domain-specialized VLMs can be compressed for fully on-device deployment without losing the safety-critical behavior required for fire detection. We develop a teacher-student knowledge distillation framework in which large VLMs fine-tuned for fire understanding can be distilled into lightweight students. Experiments across multiple VLM families and model scales show that compact students preserve most of their teachers' fire-understanding capability. We further deploy the distilled models on our commercial Detectium fire detection sensor and jointly evaluate reasoning accuracy, latency, and memory usage. The results show that compression and deployment affect not only accuracy but also model failure modes, with Qwen2.5-0.5B providing the strongest overall deployment trade-off. Our findings provide broader guidance for deploying domain-specialized VLMs in resource-constrained, safety-critical settings.
\end{abstract}


\section{Introduction}
\label{sec:intro}

Traditional fire detection systems in industrial settings rely primarily on smoke detectors or thermal cameras that trigger alarms upon detecting smoke or abnormally high temperatures~\cite{fonollosa2018chemical,khajavi2023digital}. As these sensors respond to low-level physical signals rather than the surrounding context, they suffer from high false-positive rates~\cite{liu2023fire}: benign scenarios, such as open flames during cooking or deliberately controlled high-temperature processes in a factory, can raise the same alarms as a genuinely hazardous fire~\cite{mensch2024performance}. Every false alarm requires human verification and may trigger costly interventions, wasting resources and eroding trust in the system~\cite{tannous2021economic,chagger2014causes}. Recent advances in multi-modal models, particularly vision-language models (VLMs)~\cite{liu2023visual,bai2025qwen3,team2024gemma,liu2025llava}, offer a path beyond signal-level detection. Given image frames from a camera, a VLM can reason about the scene as a whole, for example, distinguishing a controlled flame from an uncontrolled one, and assess whether an observed fire poses a genuine safety risk. Prior studies show that such semantic reasoning substantially reduces false positives compared to traditional systems~\cite{gragnaniello2025video,kim2025integrated}, pointing the way toward a more intelligent generation of fire detection systems.

Deploying VLMs for fire detection, however, is hindered by a fundamental conflict between model scale and deployment constraints. Modern VLMs with billions of parameters cannot run on the embedded hardware of fire detection sensors and must instead be hosted on cloud services, requiring every captured frame to be uploaded for analysis. This introduces two problems. The first is latency. Fire propagates rapidly, and a practical system must identify a hazardous fire at the very onset of ignition for mitigation to be effective~\cite{ccetin2013video}. Conventional smoke detectors take on the order of a minute to trigger an alarm~\cite{bukowski2003performance}. To alert before a smoke detector, a VLM-based system must therefore complete its entire detection-inference cycle within seconds. Cloud inference, however, adds round-trip communication delay, and any network failure can render the system unavailable at exactly the moment it is needed. The second, and more fundamental, is privacy. Fire sensors must be installed precisely in the spaces with the strongest privacy expectations, such as private apartments, public restrooms, or facilities where transmitting images off-site is prohibited by regulation~\cite{regulation2016679,kansal2025implications}, making continuous image upload to a cloud service unacceptable. Fire detection thus demands the semantic reasoning of large modern VLMs, yet requires fully on-device inference to guarantee low latency and privacy, \textit{requirements that current systems cannot satisfy at once}.

In this paper, we investigate how a domain-specialized VLM can be compressed sufficiently for real-world embedded deployment without sacrificing the safety-critical behavior, i.e., reduced false-alarm rate, required for fire detection. To this end, we develop a teacher-student knowledge distillation framework for on-device fire understanding. We first fine-tune state-of-the-art VLMs from the Qwen~\cite{yang2025qwen3}, Llama~\cite{grattafiori2024llama}, and Gemma~\cite{team2024gemma} families as teacher models to acquire contextual understanding of fire scenes, and then distill their fire-domain knowledge into lightweight student models spanning multiple parameter scales. We evaluate the resulting models on DetectiumFire~\cite{liu2026detectiumfire}, measuring not only their ability for fire understanding, but also their behavior on non-fire scenes that determine false-alarm performance. Our results show that compact students can retain most of their teachers' capability, while increasing student size provides surprisingly limited accuracy gains. We then deploy the distilled students on our commercial Detectium fire detection sensor and jointly evaluate their fire-understanding accuracy, safety-relevant detection errors, inference latency, and memory consumption under fully on-device inference. This deployment study reveals that model deployment involves more than an accuracy-efficiency trade-off: different architectural and deployment choices can systematically shift the model toward different failure modes. Across the tested model families and scales, Qwen2.5-0.5B provides the strongest overall deployment trade-off.

In summary, our contributions are as follows: 1. We develop a teacher-student distillation framework that compresses domain-specialized VLMs into lightweight models suitable for fully on-device fire understanding while largely preserving their semantic reasoning capability. 2. We systematically study the effect of distillation across model families and scales, showing that compact students retain most teacher performance while revealing important failure modes under different distillation objective. 3. We deploy the distilled students on our commercial Detectium fire detection sensor and characterize the practical trade-off among reasoning accuracy, safety-relevant detection errors, latency, and memory usage. Although our experiments focus on fire detection, the resulting insights on domain-specific distillation, failure-mode evaluation, and deployment-time study are broadly applicable to other safety-critical VLM applications requiring resource-constrained on-device inference.

\section{Related Work}

\subsection{VLMs for Fire Detection.}
Fire detection has traditionally been formulated as image classification or object detection~\cite{ccetin2013video,liu2023deep,elhanashi2025early,khajavi2024synthetic}, which localizes flames but cannot reason about whether a fire is hazardous. Recent work instead leverages the reasoning and language capabilities of VLMs, moving beyond detection toward semantic understanding: describing and analyzing fire scenes in natural language~\cite{seidel2025advancing}, and assessing whether an observed fire poses genuine risk~\cite{gragnaniello2025video,kim2025integrated}. Benchmarks such as SmokeBench~\cite{qi2026smokebench} and the multi-modal DetectiumFire dataset~\cite{liu2026detectiumfire} further show that VLMs can describe burning objects, environments, and risk levels beyond binary detection. However, existing work evaluates VLMs offline or assumes cloud-scale inference. Deploying fire-domain VLMs on embedded sensors remains unexplored, which is the focus of this work.

\subsection{Knowledge Distillation for VLMs.}
Knowledge distillation (KD) transfers the capability of a large teacher into a smaller student~\cite{hinton2015distilling,kim2016sequence}, and has become a standard route to deployable vision-language models~\cite{jin2025efficient,shu2025llava,cai2025llava,ghonim2026speed,kumar2026enhancing}. Several distillation frameworks have been proposed for VLMs. For example, LLaVA-MoD~\cite{shu2025llava} progressively distills a large teacher into a sparse mixture-of-experts student through mimic and preference distillation. LLaVA-KD~\cite{cai2025llava} combines output-distribution distillation with relation distillation over visual tokens in a multi-stage training scheme. Production models such as Gemma~3~\cite{team2024gemma} are likewise trained with distillation. Existing VLM distillation, however, targets general-purpose capability measured on standard multimodal benchmarks. In contrast, we study domain-specific distillation for fire understanding from a deployment perspective, evaluating not only the preservation of visual reasoning capability but also its impact on safety-critical behaviors, on-device latency, and resource consumption.

\section{Knowledge Distillation for Fire Understanding}
In this section, we present our teacher-student knowledge distillation framework for on-device fire understanding. We first describe how state-of-the-art large VLMs are fine-tuned to acquire contextual understanding of fire scenes in Section~\ref{sec:distill-teacher}. We then introduce the procedure for distilling this fire-domain knowledge into lightweight student models in Section~\ref{sec:distill-student}, following the standard distillation framework of prior work~\cite{shu2025llava}.

\subsection{Teacher Model}
\label{sec:distill-teacher}
Since general-purpose VLMs are not specifically trained to reason about fire scenes, and no existing fire-domain VLM is suitable for use as a distillation teacher, we construct our own fire-specialized teacher model $\pi_T$ through a two-stage process:

\paragraph{Foundational Initialization:} Given a pre-trained large VLM, we keep both the language model and vision encoder frozen, as their pre-trained parameters already encode rich linguistic and visual representations. We train only the vision-language adaptor, which projects visual features into the language embedding space. For this initialization stage, we use a curated image-caption dataset covering diverse scenes and visual concepts, so that the adaptor learns general-purpose visual-language alignment before any fire-specific supervision is introduced. Specifically, given a multimodal instruction example \((x,y)\), where \(x=(x_v,x_i)\) consists of the input image \(x_v\) and text instruction \(x_i\), and \(y\) denotes the ground-truth response, we optimize the standard next-token prediction objective:
\begin{equation}
\label{eq:fine-tune}
    \mathcal{L}_{\mathrm{Init}}(\pi_T)=-\mathbb{E}_{(y_k\mid y_{<k},x)\sim \pi_T}
\left[
\log \pi_T(y_k\mid y_{<k},x)
\right].
\end{equation}
Here, \(\pi_T(y_k\mid y_{<k},x)\) denotes the probability assigned by the teacher model to token \(y_k\), conditioned on the multimodal input \(x\) and the preceding response tokens \(y_{<k}=(y_1,y_2,\ldots,y_{k-1})\).

\paragraph{Task-Specific Fine-Tuning.}
The model is then fine-tuned on the fire-specific image-caption dataset to acquire contextual understanding of fire scenes. To preserve its general visual grounding during domain specialization, we additionally include general vision-language instruction data, which helps prevent the teacher from losing the ability to interpret and describe unfamiliar scenes. We optimize the model using the same next-token prediction objective as in Eq.~\ref{eq:fine-tune}.

\subsection{Student Model}
\label{sec:distill-student}

\paragraph{Foundational Initialization:}
Before distillation, we initialize the student model using the same procedure as the teacher. Specifically, we train only the vision-language adaptor on general image-caption pairs to establish general visual-language alignment before introducing fire-specific distillation.

\paragraph{Distillation.}
The initialized student model \(\pi_S\) is then trained to reproduce the teacher model's behavior on fire-understanding tasks. Given a fire-understanding example \((x,y)\) presented to both the teacher \(\pi_T\) and the student \(\pi_S\), we train the student to match the teacher's output distribution at each generated token. Following prior work~\cite{shu2025llava}, we measure the discrepancy between the teacher and student token distributions using the Kullback--Leibler (KL) divergence:
\begin{equation} 
\label{eq:KL} 
\mathcal{L}_{\mathrm{KL}}(\pi_S;\pi_T) = -\mathbb{E}_{(x,y_k)\sim \pi_T} \left[ \log \frac{ \pi_T(y_k\mid y_{<k},x) }{ \pi_S(y_k\mid y_{<k},x) } \right]
\end{equation}
Here, \(\pi_T(y_k\mid y_{<k},x)\) and \(\pi_S(y_k\mid y_{<k},x)\) denote the probabilities assigned by the teacher and student, respectively, to token \(y_k\), conditioned on the multimodal input \(x\) and the preceding tokens \(y_{<k}\). 

However, we find that optimizing the KL objective alone can cause the distilled student to predict fire for nearly every image (Table~\ref{tab:loss-ablation}), even though the corresponding teacher does not exhibit this bias. To mitigate this failure mode, we additionally incorporate a standard cross-entropy (CE) term based on the ground-truth responses, providing direct supervision from the training data:
\begin{equation}
\mathcal{L}_{\mathrm{CE}} (\pi_S) = -\mathbb{E}_{(x,y_k)\sim \pi_S}\left[
\log \pi_{\mathcal{S}}(y_t \mid x, y_{<t})\right],
\end{equation}
The student is therefore trained with the combined objective
\begin{equation}
\mathcal{L} = \mathcal{L}_{\mathrm{KL}} + \mathcal{L}_{\mathrm{CE}}.
\label{eq:combined}
\end{equation}

\section{Distillation Evaluation}
\label{sec:distill-result}

\begin{table*}[t]
\centering
\footnotesize
\setlength{\tabcolsep}{5pt}
\begin{tabular}{llcccc}
\toprule
& & \multicolumn{3}{c}{Fire-scene reasoning } & Detection \\
\cmidrule(lr){3-5}\cmidrule(lr){6-6}
Model & Vision encoder & Burning object & Environment & Severity & Fire/non-fire \\
\midrule
\multicolumn{6}{l}{\emph{Qwen2.5 family}}\\
\quad Qwen2.5-7B \textit{(Teacher)} & CLIP ViT-L/336 & 80.77 & 86.34 & 88.86 & 99.79 \\
\quad Qwen2.5-7B \textit{(Teacher)} & MobileCLIP-S2  & 81.43 & 78.51 & 83.95 & 99.72 \\
\quad Qwen2.5-0.5B & CLIP ViT-L/336 & 76.92 $\pm$ 1.16 & 85.76 $\pm$ 1.48 & 88.28 $\pm$ 1.46 & \textbf{99.75 $\pm$ 0.04} \\
\quad Qwen2.5-1.5B & CLIP ViT-L/336 & 77.19 $\pm$ 1.63 & 85.41 $\pm$ 1.61 & 87.44 $\pm$ 1.30 & 99.56 $\pm$ 0.17 \\
\quad Qwen2.5-3B   & CLIP ViT-L/336 & \textbf{78.29 $\pm$ 1.58} & 85.59 $\pm$ 0.41 & \textbf{88.82 $\pm$ 1.03} & 98.51 $\pm$ 0.46 \\
\quad Qwen2.5-3B   & MobileCLIP-S2  & 68.79 $\pm$ 1.33 & 68.61 $\pm$ 1.58 & 73.43 $\pm$ 1.22 & 96.97 $\pm$ 0.52 \\
\midrule
\multicolumn{6}{l}{\emph{Qwen3 family}}\\
\quad Qwen3-8B \textit{(Teacher)} & CLIP ViT-L/336 & 80.37 & 84.62 & 88.99 & 99.86 \\
\quad Qwen3-8B \textit{(Teacher)} & MobileCLIP-S2  & 70.69 & 84.48 & 85.54 & 99.72 \\
\quad Qwen3-0.6B & CLIP ViT-L/336 & 79.66 $\pm$ 0.17 & 80.69 $\pm$ 0.14 & \textbf{92.56 $\pm$ 0.21} & 99.68 $\pm$ 0.04 \\
\quad Qwen3-0.6B & MobileCLIP-S2  & 68.57 $\pm$ 0.48 & \textbf{86.74 $\pm$ 1.22} & 80.46 $\pm$ 0.33 & 98.74 $\pm$ 0.14 \\
\quad Qwen3-4B   & CLIP ViT-L/336 & \textbf{80.37 $\pm$ 1.48} & 86.43 $\pm$ 0.68 & 89.08 $\pm$ 0.50 & \textbf{99.61 $\pm$ 0.28} \\
\midrule
\multicolumn{6}{l}{\emph{Llama-3 family}}\\
\quad Llama-3.1-8B \textit{(Teacher)} & CLIP ViT-L/336 & 82.49 & 86.21 & 87.27 & 99.86 \\
\quad Llama-3.2-1B & CLIP ViT-L/336 & 78.16 $\pm$ 0.15 & \textbf{85.54 $\pm$ 0.13} & \textbf{87.40 $\pm$ 0.58} & \textbf{99.77 $\pm$ 0.04} \\
\quad Llama-3.2-3B & CLIP ViT-L/336 & \textbf{79.09 $\pm$ 0.63} & 84.97 $\pm$ 0.80 & 87.22 $\pm$ 1.83 & \textbf{99.77 $\pm$ 0.04} \\
\midrule
\multicolumn{6}{l}{\emph{Gemma-2 family}}\\
\quad Gemma-2-9B \textit{(Teacher)} & CLIP ViT-L/336 & 82.23 & 87.00 & 88.59 & 99.93 \\
\quad Gemma-2-2B & CLIP ViT-L/336 & 79.22 $\pm$ 1.31 & 84.84 $\pm$ 0.15 & 88.55 $\pm$ 1.25 & 99.56 $\pm$ 0.24 \\
\bottomrule
\end{tabular}
\caption{Fire-understanding accuracy (\%) across model families on the DetectiumFire evaluation set. The best-performing student within each family is shown in \textbf{bold}.
}
\label{tab:reasoning}
\vspace{-2ex}
\end{table*}

\begin{table*}[t]
\centering
\footnotesize
\setlength{\tabcolsep}{5pt}
\begin{tabular}{llccccc}
\toprule
& & \multicolumn{2}{c}{Errors (counts)} & \multicolumn{3}{c}{Rates (\%)} \\
\cmidrule(lr){3-4}\cmidrule(lr){5-7}
Model & Objective & False alarms & Missed fires & Recall & Specificity & Accuracy \\
\midrule
\multicolumn{7}{l}{\emph{Qwen2.5 family}}\\
\quad \multirow{2}{*}{Qwen2.5-0.5B}
 & $\mathcal{L}_{\mathrm{KL}}$ only & 671.3 $\pm$ 10.4 & \phantom{0}0.0 $\pm$ 0.0 & 100.00 $\pm$ 0.00 & \phantom{0}3.82 $\pm$ 1.49 & 53.76 $\pm$ 0.72 \\
 & $+\,\mathcal{L}_{\mathrm{CE}}$   & \phantom{00}2.0 $\pm$ \phantom{0}0.0 & \phantom{0}1.7 $\pm$ 0.6 & \phantom{0}99.78 $\pm$ 0.08 & 99.71 $\pm$ 0.00 & \textbf{99.75 $\pm$ 0.04} \\
\addlinespace[2pt]
\quad \multirow{2}{*}{Qwen2.5-1.5B}
 & $\mathcal{L}_{\mathrm{KL}}$ only & 549.7 $\pm$ 28.0 & \phantom{0}0.0 $\pm$ 0.0 & 100.00 $\pm$ 0.00 & 21.25 $\pm$ 4.01 & 62.14 $\pm$ 1.93 \\
 & $+\,\mathcal{L}_{\mathrm{CE}}$   & \phantom{00}4.7 $\pm$ \phantom{0}2.1 & \phantom{0}1.7 $\pm$ 0.6 & \phantom{0}99.78 $\pm$ 0.08 & 99.33 $\pm$ 0.30 & \textbf{99.56 $\pm$ 0.17} \\
\addlinespace[2pt]
\quad \multirow{2}{*}{Qwen2.5-3B}
 & $\mathcal{L}_{\mathrm{KL}}$ only & \phantom{00}4.0 $\pm$ \phantom{0}2.6 & 20.0 $\pm$ 5.3 & \phantom{0}97.35 $\pm$ 0.70 & 99.43 $\pm$ 0.38 & 98.35 $\pm$ 0.18 \\
 & $+\,\mathcal{L}_{\mathrm{CE}}$   & \phantom{0}17.3 $\pm$ \phantom{0}8.4 & \phantom{0}4.3 $\pm$ 2.5 & \phantom{0}99.43 $\pm$ 0.33 & 97.52 $\pm$ 1.20 & \textbf{98.51 $\pm$ 0.46} \\
\midrule
\multicolumn{7}{l}{\emph{Qwen3 family}}\\
\quad \multirow{2}{*}{Qwen3-4B}
 & $\mathcal{L}_{\mathrm{KL}}$ only & 275.0 $\pm$ 28.2 & \phantom{0}0.3 $\pm$ 0.6 & \phantom{0}99.96 $\pm$ 0.08 & 60.60 $\pm$ 4.03 & 81.04 $\pm$ 1.97 \\
 & $+\,\mathcal{L}_{\mathrm{CE}}$   & \phantom{00}4.3 $\pm$ \phantom{0}3.5 & \phantom{0}1.3 $\pm$ 0.6 & \phantom{0}99.82 $\pm$ 0.08 & 99.38 $\pm$ 0.50 & \textbf{99.61 $\pm$ 0.28} \\
\midrule
\multicolumn{7}{l}{\emph{Llama-3 family}}\\
\quad \multirow{2}{*}{Llama-3.2-1B}
 & $\mathcal{L}_{\mathrm{KL}}$ only & 603.7 $\pm$ \phantom{0}8.4 & \phantom{0}0.0 $\pm$ 0.0 & 100.00 $\pm$ 0.00 & 13.51 $\pm$ 1.20 & 58.43 $\pm$ 0.58 \\
 & $+\,\mathcal{L}_{\mathrm{CE}}$   & \phantom{00}0.7 $\pm$ \phantom{0}0.6 & \phantom{0}2.7 $\pm$ 0.6 & \phantom{0}99.65 $\pm$ 0.08 & 99.90 $\pm$ 0.08 & \textbf{99.77 $\pm$ 0.04} \\
\addlinespace[2pt]
\quad \multirow{2}{*}{Llama-3.2-3B}
 & $\mathcal{L}_{\mathrm{KL}}$ only & 643.0 $\pm$ 24.0 & \phantom{0}0.0 $\pm$ 0.0 & 100.00 $\pm$ 0.00 & \phantom{0}7.88 $\pm$ 3.44 & 55.72 $\pm$ 1.65 \\
 & $+\,\mathcal{L}_{\mathrm{CE}}$   & \phantom{00}1.3 $\pm$ \phantom{0}0.6 & \phantom{0}2.0 $\pm$ 0.0 & \phantom{0}99.73 $\pm$ 0.00 & 99.81 $\pm$ 0.08 & \textbf{99.77 $\pm$ 0.04} \\
\midrule
\multicolumn{7}{l}{\emph{Gemma-2 family}}\\
\quad \multirow{2}{*}{Gemma-2-2B}
 & $\mathcal{L}_{\mathrm{KL}}$ only & 226.0 $\pm$ 10.2 & \phantom{0}8.3 $\pm$ 6.4 & \phantom{0}98.89 $\pm$ 0.85 & 67.62 $\pm$ 1.45 & 83.86 $\pm$ 1.14 \\
 & $+\,\mathcal{L}_{\mathrm{CE}}$   & \phantom{00}5.0 $\pm$ \phantom{0}3.0 & \phantom{0}1.3 $\pm$ 0.6 & \phantom{0}99.82 $\pm$ 0.08 & 99.28 $\pm$ 0.43 & \textbf{99.56 $\pm$ 0.24} \\
\bottomrule
\end{tabular}
\caption{Comparison of safety-relevant detection errors under the two distillation objectives in Eq.~\ref{eq:combined} and Eq.~\ref{eq:KL}. False alarms denote non-fire images incorrectly classified as fire, while missed fires denote fire images incorrectly classified as non-fire. Recall is the proportion of fire images correctly identified, and specificity is the proportion of non-fire images correctly identified.  }
\label{tab:confusion-ablation}
\vspace{-2ex}
\end{table*}

\begin{table*}[t]
\centering
\footnotesize
\setlength{\tabcolsep}{4pt}
\resizebox{\textwidth}{!}{
\begin{tabular}{llcrcrcrcr}
\toprule
& & \multicolumn{6}{c}{Fire-scene reasoning} & \multicolumn{2}{c}{Detection} \\
\cmidrule(lr){3-8}\cmidrule(lr){9-10}
& & \multicolumn{2}{c}{Burning object} & \multicolumn{2}{c}{Environment} & \multicolumn{2}{c}{Severity} & \multicolumn{2}{c}{Fire/non-fire} \\
\cmidrule(lr){3-4}\cmidrule(lr){5-6}\cmidrule(lr){7-8}\cmidrule(lr){9-10}
Model & Vision encoder & Acc. & $\Delta$ & Acc. & $\Delta$ & Acc. & $\Delta$ & Acc. & $\Delta$ \\
\midrule
\multicolumn{10}{l}{\emph{Qwen2.5 family}}\\
\quad Qwen2.5-0.5B & CLIP ViT-L/336 & 75.95 $\pm$ 0.85 & $-0.97$ & 85.10 $\pm$ 0.43 & $-0.66$ & 88.37 $\pm$ 2.48 & $+0.09$ & 99.52 $\pm$ 0.12 & $-0.23$ \\
\quad Qwen2.5-1.5B & CLIP ViT-L/336 & 75.73 $\pm$ 1.38 & $-1.46$ & 84.13 $\pm$ 0.28 & $-1.28$ & 87.31 $\pm$ 1.20 & $-0.13$ & \textbf{99.54 $\pm$ 0.20} & $-0.02$ \\
\quad Qwen2.5-3B   & CLIP ViT-L/336 & \textbf{76.88 $\pm$ 0.41} & $-1.41$ & \textbf{86.34 $\pm$ 1.04} & $+0.75$ & \textbf{88.82 $\pm$ 0.55} & $\phantom{+}0.00$ & 97.87 $\pm$ 0.62 & $-0.64$ \\
\quad Qwen2.5-3B   & MobileCLIP-S2  & 68.39 $\pm$ 1.86 & $-0.40$ & 68.22 $\pm$ 1.89 & $-0.39$ & 73.34 $\pm$ 1.39 & $-0.09$ & 96.97 $\pm$ 0.59 & $\phantom{+}0.00$ \\
\midrule
\multicolumn{10}{l}{\emph{Qwen3 family}}\\
\quad Qwen3-0.6B & CLIP ViT-L/336 & \textbf{82.45 $\pm$ 0.20} & $+2.79$ & 76.79 $\pm$ 0.69 & $-3.90$ & 87.62 $\pm$ 0.50 & $-4.94$ & \textbf{99.38 $\pm$ 0.12} & $-0.30$ \\
\quad Qwen3-0.6B & MobileCLIP-S2  & 69.01 $\pm$ 1.00 & $+0.44$ & \textbf{87.40 $\pm$ 0.66} & $+0.66$ & 80.64 $\pm$ 0.53 & $+0.18$ & 98.81 $\pm$ 0.08 & $+0.07$ \\
\quad Qwen3-4B   & CLIP ViT-L/336 & 78.87 $\pm$ 1.23 & $-1.50$ & 85.76 $\pm$ 0.20 & $-0.67$ & \textbf{89.74 $\pm$ 0.43} & $+0.66$ & 99.29 $\pm$ 0.64 & $-0.32$ \\
\midrule
\multicolumn{10}{l}{\emph{Llama-3 family}}\\
\quad Llama-3.2-1B & CLIP ViT-L/336 & 74.09 $\pm$ 0.81 & $-4.07$ & 82.40 $\pm$ 1.10 & $-3.14$ & \textbf{86.12 $\pm$ 0.60} & $-1.28$ & \textbf{99.47 $\pm$ 0.04} & $-0.30$ \\
\quad Llama-3.2-3B & CLIP ViT-L/336 & \textbf{77.59 $\pm$ 0.27} & $-1.50$ & \textbf{84.92 $\pm$ 0.63} & $-0.05$ & 85.41 $\pm$ 2.11 & $-1.81$ & 98.88 $\pm$ 0.20 & $-0.89$ \\
\midrule
\multicolumn{10}{l}{\emph{Gemma-2 family}}\\
\quad Gemma-2-2B & CLIP ViT-L/336 & 77.59 $\pm$ 0.70 & $-1.63$ & 84.13 $\pm$ 0.80 & $-0.71$ & 88.99 $\pm$ 1.51 & $+0.44$ & 99.24 $\pm$ 0.18 & $-0.32$ \\
\bottomrule
\end{tabular}
}
\caption{Fire-understanding accuracy (\%) of the exported student models measured on the commercial Detectium fire detection sensor. \(\Delta\) denotes the change relative to the corresponding student before export, as reported in Table~\ref{tab:reasoning}. The best-performing student within each model family is shown in \textbf{bold}.}
\label{tab:cpu-accuracy}
\vspace{-2ex}
\end{table*}

\begin{table}[t]
\centering
\footnotesize
\setlength{\tabcolsep}{3.5pt}
\resizebox{\columnwidth}{!}{
\begin{tabular}{llrrrc}
\toprule
Model & Vision  & TTFT & E2E & RSS & Fits \\
& & (ms) & (ms) & (MB) & \\
\midrule
\multicolumn{6}{l}{\emph{Qwen2.5 family}}\\
\quad Qwen2.5-0.5B & CLIP       & 3818  & \textbf{4034}  & \textbf{1880} & \checkmark \\
\quad Qwen2.5-1.5B & CLIP       & 6862  & 7419  & 4210 & \checkmark \\
\quad Qwen2.5-3B   & CLIP       & 11598 & 12519 & 7341 & \checkmark \\
\quad Qwen2.5-3B   & MobileCLIP   & 1757  & 18976 & 8778 & $\times$ \\
\midrule
\multicolumn{6}{l}{\emph{Qwen3 family}}\\
\quad Qwen3-0.6B & CLIP        & 4469  & 4770  & 3050 & \checkmark \\
\quad Qwen3-0.6B & MobileCLIP   & \textbf{893}   & 4227  & 2952 & \checkmark \\
\quad Qwen3-4B   & CLIP        & 15253 & 16645 & 9850 & $\times$ \\
\midrule
\multicolumn{6}{l}{\emph{Llama-3 family}}\\
\quad Llama-3.2-1B & CLIP  & 5720  & 6905  & 3624 & \checkmark \\
\quad Llama-3.2-3B & CLIP  & 12077 & 13914 & 7548 & \checkmark \\
\midrule
\multicolumn{6}{l}{\emph{Gemma-2 family}}\\
\quad Gemma-2-2B & CLIP  & 9148 & 10146 & 7251 & \checkmark \\
\bottomrule
\end{tabular}
}
\caption{On-device latency and memory usage of the exported student models. The final column indicates whether each model fits within the memory constraints of the Detectium sensor. Models marked \(\times\) exceed the available memory and are benchmarked with additional memory for reference only. The best overall value for each metric is shown in \textbf{bold}.
}
\label{tab:latency}
\vspace{-4ex}
\end{table}

\subsection{Experimental Setup}

\paragraph{Models.}
We evaluate our distillation framework across multiple VLM families and model scales. For the teacher models, we use Qwen2.5-7B-Instruct, Qwen3-8B~\cite{yang2025qwen3}, Llama-3.1-8B-Instruct~\cite{grattafiori2024llama}, and Gemma-2-9B-IT~\cite{team2024gemma}. Each student is distilled from a teacher within the same model family. Specifically, we consider Qwen2.5-{0.5B, 1.5B, 3B}, Qwen3-{0.6B, 4B}, Llama-3.2-{1B, 3B}, and Gemma-2-2B as student models. We use the CLIP vision encoder~\cite{radford2021learning} for most configurations and additionally evaluate MobileCLIP-S2~\cite{vasu2024mobileclip} as a more lightweight alternative.

\paragraph{Training Data.}
During the foundational initialization stage, both teacher and student models are trained on the LLaVA pretraining data (LCS-558K)~\cite{liu2023visual,liu2024improved}, a 558K image-caption subset used for visual-language feature alignment. The teacher models are subsequently fine-tuned on the DetectiumFire training set~\cite{liu2026detectiumfire} to acquire contextual fire-understanding capability, together with LLaVA-1.5 Mix665K instruction-tuning data~\cite{liu2024improved} to preserve general visual-language capability during domain specialization. The student models are then distilled using only the DetectiumFire training set.

\paragraph{Evaluation.}
We evaluate all teacher and student models on the DetectiumFire evaluation set, which contains 1{,}452 images, including 754 fire images and 698 non-fire images. We consider four fire-understanding tasks: fire/non-fire detection, identification of the burning object, recognition of the surrounding environment, and assessment of fire severity. The latter three tasks are evaluated on fire images only, while fire/non-fire detection is evaluated on the complete evaluation set. All results for student models are averaged over three random seeds.

Additional implementation details are provided in Appendix~\ref{app:implement}.

\subsection{Results on Fire Understanding}

Table~\ref{tab:reasoning} reports accuracy on the four fire-understanding tasks for each teacher and its distilled students. Overall, the teacher models achieve consistently strong performance across model families, despite differences in backbone and parameter scale. In particular, their near-perfect fire/non-fire detection accuracy shows that fine-tuned VLMs can effectively distinguish hazardous fire scenes from benign ones, while the more fine-grained fire-scene reasoning tasks remain more challenging.

\emph{Students retain most of the teacher's capability across every family.} For example, relative to its teacher, Qwen2.5-0.5B retains 95.2\% of the burning-object accuracy, 99.3\% of the environment and severity accuracy, and 99.96\% of the fire/non-fire detection accuracy, while using roughly an order of magnitude fewer language-model parameters (0.5B vs.\ 7B). Similar patterns hold across the other model families, demonstrating that the distillation framework preserves most of the fire-understanding capability of the teacher models. However, \emph{the distillation gap is not uniform across tasks, with the largest degradation occurring in burning-object identification}. Across families, students typically trail their teachers by approximately 2.5-4.3 on this task, whereas the gap in fire/non-fire detection is negligible. Environment and severity prediction fall between these two extremes, with several students matching or even exceeding their teachers. This suggests that fine-grained visual discrimination, such as identifying the specific object that is burning, is more sensitive to distillation than coarse fire detection or scene-level reasoning.

Within the student models, \emph{increasing model capacity provides little additional accuracy}. For example, scaling Qwen2.5 from 0.5B to 3B improves burning-object accuracy by only 1.37, while the corresponding gains for the Llama-3 and Qwen3 families are 0.93 and 0.71, respectively, all within or close to one standard deviation across seeds. For fire/non-fire detection, larger students provide no benefit: Qwen2.5 accuracy decreases monotonically from 99.75\% at 0.5B to 99.56\% at 1.5B and 98.51\% at 3B, and no model family shows a consistent improvement with increasing student size. Since latency and memory usage grow substantially with parameter count (Section~\ref{sec:deployment}, Table~\ref{tab:latency}), \emph{the smallest student is not merely a deployment compromise, but is often the preferred choice in terms of both accuracy and efficiency}.

Finally, \emph{the choice of vision encoder has a larger effect on accuracy than increasing the language-model size}. Replacing CLIP ViT-L/336 with MobileCLIP-S2 causes substantially larger performance drops than those observed when changing student scale. The effect is already visible in the teachers: Qwen2.5-7B loses 7.83 in environment accuracy and 4.91 in severity, while Qwen3-8B loses 9.68 in burning-object accuracy. This degradation is subsequently inherited by the distilled students. Therefore, although a lightweight vision encoder can reduce computational cost (Section~\ref{sec:deployment}, Table~\ref{tab:latency}), the efficiency gains discussed in Section~\ref{sec:deployment} come at a substantial loss in fire-understanding accuracy.

\subsection{Ablation Study}

Table~\ref{tab:confusion-ablation} isolates the effect of the distillation objective by comparing students trained with the full objective in Eq.~\ref{eq:combined} against those trained with the KL term in Eq.~\ref{eq:KL} alone. \emph{Removing the cross-entropy term \(\mathcal{L}_{\mathrm{CE}}\) causes a severe collapse in fire/non-fire detection.} In particular, Qwen2.5-0.5B, Qwen2.5-1.5B, Llama-3.2-1B, and Llama-3.2-3B all achieve \(100.00 \pm 0.00\%\) recall but only 3.82\%-21.25\% specificity, indicating that they classify nearly every image as containing fire. In contrast, adding \(\mathcal{L}_{\mathrm{CE}}\) restores balanced detection performance across all students, with specificity between 97.52\% and 99.90\% and recall between 99.43\% and 99.82\%.

\emph{This collapse is not mitigated by increasing model capacity.} Qwen3-4B, the largest student considered in this ablation, still achieves only 60.60\% specificity without the cross-entropy term, while Llama-3.2-3B performs even worse than the smaller Llama-3.2-1B (7.88\% vs.\ 13.51\% specificity). We further report teacher-model results in Appendix~\ref{app:experiment-result}, Table~\ref{tab:confusion-teachers}, where every teacher produces at most two false alarms. Thus, \emph{the collapse is not inherited from the teacher}, but instead emerges during distillation.

The complete ablation results on the fire-understanding tasks are reported in Appendix~\ref{app:experiment-result}, Table~\ref{tab:loss-ablation}. Strikingly, \emph{this failure is largely invisible on the fire-scene reasoning tasks}: removing \(\mathcal{L}_{\mathrm{CE}}\) leaves burning-object, environment, and severity accuracy largely unchanged, and in several cases even improves them. Because these tasks are evaluated only on images containing fire, a model that predicts fire indiscriminately is not penalized. A practitioner selecting the distillation objective based only on positive-class reasoning performance could therefore favor the KL-only objective despite its severe deployment failure. \textbf{This result highlights a broader lesson: domain-specific distillation must be validated on negative examples, not only on the positive class that defines the task.}

\section{On-device Deployment}
\label{sec:deployment}

\subsection{Deployment Setup}

To evaluate the real-world performance of the distilled students, we deploy them on a commercial Detectium fire detection sensor used for industrial fire monitoring~\footnote{\url{https://www.detectium.io/sensor-package}}~\cite{khajavi2023digital}. The target sensor has limited memory and contains no GPU or dedicated hardware accelerator. Moreover, it operates without network access during inference, so all experiments are performed entirely on the sensor CPU. This setting reflects the resource and privacy constraints of real-world deployment discussed in Section~\ref{sec:intro}.

For embedded inference, student models using the CLIP vision encoder are exported with the \texttt{llama.cpp} toolchain~\footnote{\url{https://github.com/ggml-org/llama.cpp}}. The language-model weights are converted to GGUF format and quantized to Q8\_0, while the vision encoder and vision-language adaptor are packaged separately as the multimodal component. Because \texttt{llama.cpp} does not support conversion of the MobileCLIP-S2 encoder, students using MobileCLIP-S2 are instead executed directly on the device without this export pipeline.

We evaluate deployment efficiency using three metrics. \emph{Time to first token (TTFT)} measures the latency from receiving an input request to generating the first output token. \emph{End-to-end latency (E2E)} measures the total inference time, including both prefill and response generation. \emph{Peak resident set size (RSS)} measures the maximum physical memory occupied by the inference process. In addition to these efficiency metrics, we re-evaluate the deployed models on the same DetectiumFire evaluation set to measure how deployment affects their fire-understanding capability and detection behavior. Additional hardware, export, and measurement details are provided in Appendix~\ref{app:deployment}.

\subsection{Results}

Table~\ref{tab:cpu-accuracy} reports the accuracy of the exported students measured on the Detectium sensor, together with the change relative to the corresponding models before export in Table~\ref{tab:reasoning}. Overall, \emph{fire/non-fire detection accuracy is largely preserved after deployment}, with an absolute change of at most 0.89 percentage across all students. In contrast, \emph{fine-grained fire-scene reasoning is more sensitive to deployment-time compression}. For example, Llama-3.2-1B loses 4.07 in burning-object accuracy and 3.14 in environment accuracy after export. The two MobileCLIP-based students, which are executed directly without conversion or quantization, exhibit much smaller changes, with a maximum absolute difference of only 0.66 across the four tasks. This suggests that the degradation observed for the CLIP-based students is primarily associated with the export and quantization pipeline rather than with moving inference from GPU to CPU alone. Consistent with the distillation results in Section~\ref{sec:distill-result}, fine-grained visual discrimination is the capability most sensitive to compression: it degrades under both distillation and deployment-time quantization. Nevertheless, the overall loss remains modest. For example, after both distillation and quantization, Qwen2.5-0.5B remains only 5.88 below its 7B teacher in burning-object accuracy.

To better understand changes in detection behavior, Appendix~\ref{app:deploy-result}, Table~\ref{tab:cpu-confusion}, decomposes fire/non-fire errors into false alarms and missed fires. \emph{For the quantized CLIP-based students, deployment introduces a directional shift toward predicting fire more frequently}, increasing false alarms rather than missed fires. Moreover, \emph{this shift tends to become more pronounced for larger students}, suggesting that smaller students are not only faster and more memory-efficient, but can also be more robust to deployment-time compression. MobileCLIP-based students exhibit the opposite and more safety-critical behavior: \emph{their errors are more likely to take the form of missed fires}. In practical fire detection, a missed fire is substantially more consequential than a false alarm, making this failure mode undesirable despite the computational efficiency of the lightweight vision encoder.

Table~\ref{tab:latency} reports on-device latency and memory usage for each student. We find that \emph{latency is dominated by prefill rather than autoregressive generation}. For every CLIP-based student, TTFT accounts for approximately 83\%-95\% of the total E2E latency. For example, Qwen2.5-0.5B requires 3818 ms to produce its first token but only an additional 216 ms to generate the remainder of the response. 
We further find that \emph{the vision encoder is a major determinant of prefill latency}. Replacing CLIP with MobileCLIP-S2 reduces TTFT substantially: from 4469 ms to 893 ms for Qwen3-0.6B and from 11598 ms to 1757 ms for Qwen2.5-3B. This reduction is consistent with the substantially smaller visual representation produced by MobileCLIP-S2. However, because MobileCLIP-S2 is not supported by the \texttt{llama.cpp} export pipeline and must instead be executed directly on the device, these models do not obtain the same optimized inference path. Consequently, the reduction in TTFT does not necessarily translate into lower E2E latency.

Finally, \emph{deployment cost grows steeply with student size, while accuracy does not}. Across the evaluated Qwen models, E2E latency increases from 4034 ms for Qwen2.5-0.5B to 16645 ms for Qwen3-4B, while RSS increases from 1880 MB to 9850 MB. In contrast, the improvement in fire-scene reasoning accuracy from using larger students is limited, and the smallest models remain highly competitive in fire/non-fire detection (Table~\ref{tab:cpu-accuracy}). Considering fire-understanding capability, detection reliability, latency, and memory jointly, \textbf{Qwen2.5-0.5B provides the strongest overall deployment trade-off}: it is the fastest configuration at 4034 ms E2E latency, approximately an order of magnitude lower than that reported for conventional smoke detectors~\cite{bukowski2003performance}. It also has the lowest memory usage and achieves overall best fire-understanding accuracy after export.

\bibliography{aaai2027}

\clearpage
\appendix

\section{Conclusion}
In this paper, we studied how far domain-specialized VLMs can be compressed for fully on-device fire understanding without sacrificing the behaviors required for reliable deployment. Across multiple VLM families and model scales, our experiments show that compact students preserve most of their teachers' fire-understanding capability, while increasing student size provides only limited additional benefit. At the same time, our ablation study reveals that distillation quality cannot be judged solely by positive-class reasoning accuracy: distillation objective with only the KL term as in Eq.~\ref{eq:KL} can preserve strong fire-scene reasoning while inducing severe false-alarm bias, highlighting the importance of negative examples in evaluating domain-specific compression. Real-world deployment further shows that compression affects not only accuracy, but also the type of errors made by the system, with different architectural and quantization choices shifting the balance between false alarms and missed fires. Among the evaluated models, Qwen2.5-0.5B provides the strongest overall trade-off between reasoning capability, detection reliability, latency, and memory usage on the target sensor.

Although our experiments focus on fire detection, these findings are not specific to this domain. Other safety-critical and resource-constrained applications of VLMs face the same challenge of preserving domain-specific reasoning while satisfying strict deployment constraints. Our results therefore suggest broader lessons for on-device VLM deployment: smaller models can be preferable to larger ones, negative examples are essential for validating distilled models, and deployment-time compression should be evaluated in terms of failure modes as well as accuracy and efficiency.

\section{Additional Implementation Details}
\label{app:implement}

\subsection{Model Architecture}

All teacher and student models follow the LLaVA-MoD architecture~\cite{shu2025llava}, which consists of three main components: a vision encoder, a vision-language (VL) adaptor, and a large language model (LLM). We summarize the architecture here for completeness. Given a multimodal instruction pair \(x=(x_v,x_i)\), where \(x_v\) denotes the input image and \(x_i\) the text instruction, the model generates response \(y\) as
\[
y = \mathrm{LLM}\bigl(\mathrm{Proj}(\mathrm{ViT}(x_v)), x_i\bigr).
\]

\(\mathrm{ViT}\) denotes the vision encoder, which extracts visual features from \(x_v\). We consider two vision encoders in our experiments. The first is the CLIP vision encoder (\texttt{openai/clip-vit-large-patch14-336})~\cite{radford2021learning}, which takes a \(336\times336\) input image and produces \(576\) visual tokens using \(14\times14\) image patches. The second is MobileCLIP-S2 (MCi2)~\cite{vasu2024mobileclip}, a lightweight vision encoder based on FastViT~\cite{vasu2023fastvit}, which takes a \(256\times256\) input image and produces 64 visual tokens. We load the \texttt{apple/MobileCLIP-S2-OpenCLIP} checkpoint through OpenCLIP~\cite{ilharco2021openclip}.

\(\mathrm{Proj}\) denotes the VL adaptor, which maps the visual features produced by the vision encoder into the language-model embedding space. Following LLaVA-MoD~\cite{shu2025llava}, we implement the adaptor as a two-layer MLP with a GELU activation.

\(\mathrm{LLM}\) denotes the language-model backbone, which generates the response \(y\) conditioned on the projected visual features and text instruction. For the teacher models, we consider Qwen2.5-7B-Instruct, Qwen3-8B~\cite{yang2025qwen3}, Llama-3.1-8B-Instruct~\cite{grattafiori2024llama}, and Gemma-2-9B-IT~\cite{team2024gemma}. Following LLaVA-MoD~\cite{shu2025llava}, each student is distilled from a teacher with a compatible vocabulary. Specifically, Qwen2.5-7B is distilled into Qwen2.5-{0.5B, 1.5B, 3B}-Instruct students; Qwen3-8B is distilled into Qwen3-{0.6B, 4B} students; Llama-3.1-8B is distilled into Llama-3.2-{1B, 3B}-Instruct students; and Gemma-2-9B is distilled into Gemma-2-2B-IT. Although the Llama teacher and students belong to different Llama releases, Llama-3.1 and Llama-3.2 use the same tokenizer and therefore share the vocabulary required for token-level distillation. Finally, unlike LLaVA-MoD~\cite{shu2025llava}, we do not sparsify the student LLMs. Each student retains the standard architecture of its corresponding base language model.

\subsection{Training Data}

\paragraph{Teacher Model.}

For the foundational initialization stage, we train the vision-language adaptor on the \texttt{liuhaotian/LLaVA-Pretrain} dataset~\cite{liu2023visual,liu2024improved}, a filtered set of 558K image-caption pairs used for visual-language feature alignment. For the subsequent task-specific fine-tuning stage, we use 35,486 training examples, whose composition is summarized in Table~\ref{tab:data-teacher}.

\begin{table}[h]
\centering\small
\begin{tabular}{lrl}
\toprule
Subset & NO. & Source \\
\midrule
Description-style  & 12{,}000 & LLaVA mix665k \\
Localization  & 2{,}409  & LLaVA mix665k \\
Simple VQA          & 3{,}000  & LLaVA mix665k \\
Fire VQA   & 13{,}077 & DetectiumFire train \\
Bbox + non-fire     & 5{,}000  & DetectiumFire train \\
\bottomrule
\end{tabular}
\caption{Composition of the training data used in the task-specific fine-tuning stage.}
\label{tab:data-teacher}
\vspace{-2ex}
\end{table}

The first three subsets are used to preserve the teacher models' general visual-language capability during fire-specific fine-tuning and are constructed by filtering the LLaVA-1.5 Mix665K dataset. We retain single-turn English examples associated with COCO train2017 or Visual Genome images and group them according to instruction type. Description-style examples contain image-description prompts; localization examples request bounding-box coordinates and are restricted to COCO images; and simple VQA examples contain short factual or multiple-choice questions after excluding description, localization, and multi-step reasoning prompts.

To equip the teacher model with contextual fire-understanding capability, we fine-tune it on the DetectiumFire training set~\cite{liu2026detectiumfire}, a comprehensive multimodal fire dataset containing visual question-answering annotations for burning objects, surrounding environments, and fire severity, together with non-fire images. We use the official training split for fine-tuning and reserve the validation split for evaluation. For fire images, we directly use the provided visual question-answering (VQA) annotations as supervision. We additionally include 3{,}000 real-world non-fire images from DetectiumFire. This negative supervision is particularly important because false alarms are a central challenge in practical fire detection (Section~\ref{sec:intro}). It encourages the teacher not only to characterize fire scenes, but also to correctly reject benign scenes that may superficially resemble fire.

We additionally include 2,000 synthetic fire images with bounding-box annotations and 3,000 real non-fire images as negative localization examples. Both subsets use the instruction, “Please provide all bounding box coordinates for regions with visible fire or flame.” For synthetic fire images, the target response consists of the annotated boxes converted to normalized \(xyxy\) coordinates. For non-fire images, the response is "No fire or flame is visible in this image. There are no bounding boxes to report." These examples are used only during teacher fine-tuning and are separate from the DetectiumFire VQA supervision.

\paragraph{Student Model.}
For the foundational initialization stage, following the same procedure as for the teacher model, we train the vision-language adaptor using the \texttt{liuhaotian/LLaVA-Pretrain} dataset~\cite{liu2023visual,liu2024improved}. For the subsequent distillation stage, we use only the DetectiumFire training set.

To expose the student to diverse formulations of the fire-understanding task, each training image is paired separately with multiple prompts. Specifically, each fire image is used four times, once with each of the following prompts: (1) ``Carefully analyze the image for signs of fire or smoke.'' (2) ``What is burning, where is it happening, and how severe is the fire?'' (3) ``Describe the fire scene and estimate risk level.'' and (4) ``Provide a concise fire-focused description of this image.'' Similarly, each non-fire image is used three times, once with each of the following prompts: (5) ``Carefully analyze the image for signs of fire or smoke.'' (6) ``Check this scene for signs of fire, flame, or smoke.'' and (7) ``Is there any evidence of active fire in this image? Explain briefly.'' Thus, each image-prompt pair constitutes a separate distillation example.

During evaluation, we use these same task formulations and additionally introduce the prompt ``Assess whether this is a fire incident or a non-fire scene.'' This additional formulation is not used during student distillation and therefore provides an evaluation of generalization to unseen question wording.

\subsection{Training Configuration}

All models are trained on eight NVIDIA A100 GPUs with 80,GB of memory per GPU.

\paragraph{Teacher Model.}
We fine-tune the teacher models using low-rank adaptation (LoRA)~\cite{hu2021lora} rather than full-parameter fine-tuning. The detailed training hyperparameters are summarized in Table~\ref{tab:training-hyperparameters-teacher}.

\begin{table}[t]
\centering
\caption{Training hyperparameters for teacher initialization and task-specific fine-tuning.}
\label{tab:training-hyperparameters-teacher}
\vspace{-1ex}
\resizebox{\columnwidth}{!}{
\begin{tabular}{lcc}
\toprule
\textbf{Configuration}
& \textbf{Initialization}
& \textbf{Fine-Tuning}
\\
\midrule
LLM
& \ding{55}
& \ding{51} (LoRA)
\\
VL Adaptor
& \ding{51}
& \ding{51}
\\
ViT
& \ding{55}
& \ding{55}
\\
\midrule
LLM sequence length
& 2048
& 1536
\\
Optimizer
& AdamW
& AdamW
\\
Optimizer hyperparameter
& \(\beta_1=0.9,\ \beta_2=0.999\)
& \(\beta_1=0.9,\ \beta_2=0.999\)
\\
Learning rate
& \(1\times10^{-3}\) (adaptor)
& \(2\times10^{-4}\) (LoRA), \(2\times10^{-5}\) (adaptor)
\\
LoRA rank \(r\)
& --
& 128
\\
LoRA \(\alpha\)
& --
& 256
\\
LoRA dropout
& --
& 0.05
\\
Learning rate schedule
& cosine
& cosine
\\
Weight decay
& 0
& 0
\\
Training epoch
& 1
& 1
\\
Warm-up ratio
& 0.03
& 0.03
\\
Global batch size
& 128
& 64
\\
Numerical precision
& bf16
& bf16
\\
Model parallelism
& DeepSpeed ZeRO-2
& DeepSpeed ZeRO-2
\\
\bottomrule
\end{tabular}
}
\vspace{-1ex}
\end{table}

\paragraph{Student Model.}

The detailed training hyperparameters for the student models are summarized in Table~\ref{tab:training-hyperparameters-student}. All experiments are conducted using the same three random seeds: 42, 123, and 456.

\begin{table}[t]
\centering
\caption{Training hyperparameters for student initialization and distillation.}
\label{tab:training-hyperparameters-student}
\vspace{-1ex}
\resizebox{\columnwidth}{!}{
\begin{tabular}{lcc}
\toprule
\textbf{Configuration}
& \textbf{Initialization}
& \textbf{Distillation}
\\
\midrule
LLM
& \ding{55}
& \ding{51} (LoRA)
\\
VL Adaptor
& \ding{51}
& \ding{51}
\\
ViT
& \ding{55}
& \ding{55}
\\
\midrule
LLM sequence length
& 2048
& 2048
\\
Optimizer
& AdamW
& AdamW
\\
Optimizer hyperparameter
& \(\beta_1=0.9,\ \beta_2=0.999\)
& \(\beta_1=0.9,\ \beta_2=0.999\)
\\
Learning rate
& \(1\times10^{-3}\)
& \(2\times10^{-5}\)
\\
LoRA rank \(r\)
& --
& 128
\\
LoRA \(\alpha\)
& --
& 256
\\
LoRA dropout
& --
& 0.05
\\
Learning rate schedule
& cosine
& cosine
\\
Weight decay
& 0
& 0
\\
Training epoch
& 1
& 1
\\
Warm-up ratio
& 0.03
& 0.03
\\
Global batch size
& 128
& 64
\\
Numerical precision
& bf16
& bf16
\\
Model parallelism
& DeepSpeed ZeRO-2
& DeepSpeed ZeRO-2 / ZeRO-2 offload
\\
\bottomrule
\end{tabular}
}
\vspace{-2ex}
\end{table}

\section{Additional Experimental Results}
\label{app:experiment-result}

\subsection{Ablation Study}

\begin{table*}[t]
\centering
\footnotesize
\setlength{\tabcolsep}{5pt}
\begin{tabular}{llccccc}
\toprule
& & \multicolumn{3}{c}{Fire-scene reasoning} & \multicolumn{2}{c}{Detection } \\
\cmidrule(lr){3-5}\cmidrule(lr){6-7}
Model & Objective & Burning object & Environment & Severity & Fire/non-fire & $\Delta$ \\
\midrule
\multicolumn{7}{l}{\emph{Qwen2.5 family}}\\
\quad \multirow{2}{*}{Qwen2.5-0.5B}
 & $\mathcal{L}_{\mathrm{KL}}$ only & 77.59 $\pm$ 0.48 & 86.16 $\pm$ 0.28 & 88.81 $\pm$ 0.31 & 53.76 $\pm$ 0.72 & \\
 & $+\,\mathcal{L}_{\mathrm{CE}}$   & 76.92 $\pm$ 1.16 & 85.76 $\pm$ 1.48 & 88.28 $\pm$ 1.46 & \textbf{99.75 $\pm$ 0.04} & \textbf{+45.99} \\
\addlinespace[2pt]
\quad \multirow{2}{*}{Qwen2.5-1.5B}
 & $\mathcal{L}_{\mathrm{KL}}$ only & 79.62 $\pm$ 0.15 & 87.23 $\pm$ 0.43 & 88.46 $\pm$ 0.27 & 62.14 $\pm$ 1.93 & \\
 & $+\,\mathcal{L}_{\mathrm{CE}}$   & 77.19 $\pm$ 1.63 & 85.41 $\pm$ 1.61 & 87.44 $\pm$ 1.30 & \textbf{99.56 $\pm$ 0.17} & \textbf{+37.42} \\
\addlinespace[2pt]
\quad \multirow{2}{*}{Qwen2.5-3B}
 & $\mathcal{L}_{\mathrm{KL}}$ only & 76.44 $\pm$ 1.26 & 85.23 $\pm$ 0.81 & 86.47 $\pm$ 0.87 & 98.35 $\pm$ 0.18 & \\
 & $+\,\mathcal{L}_{\mathrm{CE}}$   & 78.29 $\pm$ 1.58 & 85.59 $\pm$ 0.41 & 88.82 $\pm$ 1.03 & \textbf{98.51 $\pm$ 0.46} & +0.16 \\
\midrule
\multicolumn{7}{l}{\emph{Qwen3 family}}\\
\quad \multirow{2}{*}{Qwen3-4B}
 & $\mathcal{L}_{\mathrm{KL}}$ only & 80.02 $\pm$ 0.31 & 86.21 $\pm$ 0.48 & 88.73 $\pm$ 0.58 & 81.04 $\pm$ 1.97 & \\
 & $+\,\mathcal{L}_{\mathrm{CE}}$   & 80.37 $\pm$ 1.48 & 86.43 $\pm$ 0.68 & 89.08 $\pm$ 0.50 & \textbf{99.61 $\pm$ 0.28} & \textbf{+18.57} \\
\midrule
\multicolumn{7}{l}{\emph{Llama-3 family}}\\
\quad \multirow{2}{*}{Llama-3.2-1B}
 & $\mathcal{L}_{\mathrm{KL}}$ only & 81.21 $\pm$ 0.50 & 87.97 $\pm$ 0.50 & 88.11 $\pm$ 0.63 & 58.43 $\pm$ 0.58 & \\
 & $+\,\mathcal{L}_{\mathrm{CE}}$   & 78.16 $\pm$ 0.15 & 85.54 $\pm$ 0.13 & 87.40 $\pm$ 0.58 & \textbf{99.77 $\pm$ 0.04} & \textbf{+41.34} \\
\addlinespace[2pt]
\quad \multirow{2}{*}{Llama-3.2-3B}
 & $\mathcal{L}_{\mathrm{KL}}$ only & 79.44 $\pm$ 0.40 & 86.61 $\pm$ 0.61 & 86.69 $\pm$ 0.67 & 55.72 $\pm$ 1.65 & \\
 & $+\,\mathcal{L}_{\mathrm{CE}}$   & 79.09 $\pm$ 0.63 & 84.97 $\pm$ 0.80 & 87.22 $\pm$ 1.83 & \textbf{99.77 $\pm$ 0.04} & \textbf{+44.05} \\
\midrule
\multicolumn{7}{l}{\emph{Gemma-2 family}}\\
\quad \multirow{2}{*}{Gemma-2-2B}
 & $\mathcal{L}_{\mathrm{KL}}$ only & 79.93 $\pm$ 0.80 & 86.61 $\pm$ 0.93 & 79.93 $\pm$ 2.28 & 83.86 $\pm$ 1.14 & \\
 & $+\,\mathcal{L}_{\mathrm{CE}}$   & 79.22 $\pm$ 1.31 & 84.84 $\pm$ 0.15 & 88.55 $\pm$ 1.25 & \textbf{99.56 $\pm$ 0.24} & \textbf{+15.70} \\
\bottomrule
\end{tabular}
\caption{Effect of the distillation objective on fire-understanding accuracy (\%) across model families on the DetectiumFire evaluation set. \(\Delta\) denotes the change in fire/non-fire detection accuracy. Removing \(\mathcal{L}_{\mathrm{CE}}\) leaves performance on the three fire-scene reasoning tasks largely unchanged, and in several cases slightly improves it, while reducing fire/non-fire detection accuracy by up to 45.99 percentage.
}
\vspace{-2ex}
\label{tab:loss-ablation}
\end{table*}

Table~\ref{tab:loss-ablation} compares students trained with the full objective in Eq.~\ref{eq:combined} against those trained with the KL term in Eq.~\ref{eq:KL} alone. \emph{Removing \(\mathcal{L}_{\mathrm{CE}}\) causes a severe collapse in fire/non-fire detection}, with most students losing between 15.70 and 45.99 percentage points in accuracy. This degradation persists across model scales. For example, Qwen2.5-0.5B drops to 53.76\% accuracy, while Qwen3-4B, despite being eight times larger, still falls to 81.04\%. Similarly, Llama-3.2-3B performs even worse than the smaller Llama-3.2-1B (55.72\% vs.\ 58.43\%).

In contrast, \emph{the collapse is nearly invisible on the fire-scene reasoning tasks}. Removing the cross-entropy term \(\mathcal{L}_{\mathrm{CE}}\) leaves burning-object, environment, and severity accuracy largely unchanged and, in many cases, even nominally improves performance. For example, burning-object accuracy increases for four of the seven students, including gains of 3.05 points for Llama-3.2-1B and 2.43 points for Qwen2.5-1.5B. Because these three tasks are evaluated only on images containing fire, a model that predicts fire indiscriminately is not penalized by these benchmarks. Consequently, a practitioner selecting a distillation objective solely on the basis of fire-scene reasoning performance could favor the KL-only objective even though it produces a model that is unsuitable for deployment. This ablation therefore highlights a central practical lesson: \textbf{domain-specific distillation must be validated on negative examples.}

\begin{table}[t]
\centering
\small
\setlength{\tabcolsep}{5pt}
\resizebox{\columnwidth}{!}{%
\begin{tabular}{llccccc}
\toprule
& & \multicolumn{2}{c}{Errors (counts)} & \multicolumn{3}{c}{Rates (\%)} \\
\cmidrule(lr){3-4}\cmidrule(lr){5-7}
Teacher & Encoder & False alarms & Missed fires & Recall & Specificity & Accuracy \\
\midrule
Qwen2.5-7B           & CLIP       & 2 & 1 & 99.87 & 99.71 & 99.79 \\
Qwen2.5-7B           & MobileCLIP & 2 & 2 & 99.73 & 99.71 & 99.72 \\
Qwen3-8B             & CLIP       & 1 & 1 & 99.87 & 99.86 & 99.86 \\
Qwen3-8B             & MobileCLIP & 2 & 2 & 99.73 & 99.71 & 99.72 \\
Llama-3.1-8B         & CLIP       & 1 & 1 & 99.87 & 99.86 & 99.86 \\
Gemma-2-9B           & CLIP       & 1 & 0 & 100.00 & 99.86 & 99.93 \\
\bottomrule
\end{tabular}
}
\caption{Safety-relevant detection errors for the teacher models. }
\label{tab:confusion-teachers}
\vspace{-4ex}
\end{table}

Table~\ref{tab:confusion-teachers} reports fire/non-fire detection performance for the teacher models. All teachers achieve consistently strong results, producing at most two false alarms and at most two missed fires. This confirms that the collapse observed in students trained without \(\mathcal{L}_{\mathrm{CE}}\) (Table~\ref{tab:confusion-ablation}) is not inherited from the teacher models, but instead emerges during distillation.

\section{Additional Deployment Configuration}
\label{app:deployment}

\subsection{Target Sensor}

We evaluate the distilled students on our commercial Detectium fire detection sensor~\footnote{\url{https://www.detectium.io/sensor-package}}. The sensor is built on a Raspberry Pi 5 with 8 GB of memory and runs Raspberry Pi OS Lite 64-bit (Debian 13, glibc 2.41). It is equipped with a Broadcom BCM2712 SoC containing four Arm Cortex-A76 CPU cores (ARMv8.2-A) clocked at up to 2.4 GHz. No GPU or NPU is used during inference. To reserve sufficient memory for the operating system and other system processes, we impose a 7 GiB memory cap on model inference. All experiments use four CPU threads, one per core.

\begin{table*}[h]
\centering\footnotesize
\begin{tabular}{lp{0.78\textwidth}}
\toprule
\textbf{Family} & \textbf{Prompt template} \\
\midrule

Qwen2.5 &
\texttt{USER: <image>\{prompt\} ASSISTANT:}
\\[1ex]

Qwen3 &
\begin{minipage}[t]{0.78\textwidth}
\ttfamily
<|im\_start|>system\\
\{system\}<|im\_end|>\\
<|im\_start|>user\\
<image>\\
\{prompt\}<|im\_end|>\\
<|im\_start|>assistant
\end{minipage}
\\[2ex]

Llama-3.2 &
\begin{minipage}[t]{0.78\textwidth}
\ttfamily
<|begin\_of\_text|><|start\_header\_id|>system<|end\_header\_id|>\\
\\
\{system\}<|eot\_id|><|start\_header\_id|>user<|end\_header\_id|>\\
\\
<image>\\
\{prompt\}<|eot\_id|><|start\_header\_id|>assistant<|end\_header\_id|>
\end{minipage}
\\[2ex]

Gemma-2 &
\begin{minipage}[t]{0.78\textwidth}
\ttfamily
<start\_of\_turn>user\\
<image>\\
\{prompt\}<end\_of\_turn>\\
<start\_of\_turn>model
\end{minipage}
\\

\bottomrule
\end{tabular}
\caption{Prompt templates used for each student family at inference, matching the formats used during training. For Qwen3 and Llama-3.2, \texttt{\{system\}} denotes the system message specified in the text. Gemma-2 uses an empty system prompt.}
\label{tab:prompt-template}
\vspace{-2ex}
\end{table*}

\begin{table*}[t]
\centering
\footnotesize
\setlength{\tabcolsep}{4pt}
\resizebox{\textwidth}{!}{
\begin{tabular}{llcrcrcrcr}
\toprule
& & \multicolumn{4}{c}{Errors (counts)} & \multicolumn{4}{c}{Rates (\%)} \\
\cmidrule(lr){3-6}\cmidrule(lr){7-10}
& & \multicolumn{2}{c}{False alarms} & \multicolumn{2}{c}{Missed fires} & \multicolumn{2}{c}{Recall} & \multicolumn{2}{c}{Specificity} \\
\cmidrule(lr){3-4}\cmidrule(lr){5-6}\cmidrule(lr){7-8}\cmidrule(lr){9-10}
Model & Vision encoder & Count & $\Delta$ & Count & $\Delta$ & Rate & $\Delta$ & Rate & $\Delta$ \\
\midrule
\multicolumn{10}{l}{\emph{Qwen2.5 family}}\\
\quad Qwen2.5-0.5B & CLIP ViT-L/336 & \phantom{0}5.7 $\pm$ 2.5 & $+3.7$ & \phantom{0}1.3 $\pm$ 1.2 & $-0.4$ & 99.83 $\pm$ 0.16 & $+0.05$ & 99.18 $\pm$ 0.36 & $-0.53$ \\
\quad Qwen2.5-1.5B & CLIP ViT-L/336 & \phantom{0}5.0 $\pm$ 1.7 & $+0.3$ & \phantom{0}1.7 $\pm$ 1.2 & $\phantom{+}0.0$ & 99.77 $\pm$ 0.16 & $\phantom{+}0.00$ & 99.28 $\pm$ 0.24 & $-0.04$ \\
\quad Qwen2.5-3B   & CLIP ViT-L/336 & 27.7 $\pm$ 9.0 & $+10.4$ & \phantom{0}3.3 $\pm$ 0.6 & $-1.0$ & 99.56 $\pm$ 0.08 & $+0.13$ & 96.03 $\pm$ 1.29 & $-1.49$ \\
\quad Qwen2.5-3B   & MobileCLIP-S2  & \phantom{0}4.7 $\pm$ 1.5 & $+0.4$ & 39.3 $\pm$ 9.7 & $-0.4$ & 94.79 $\pm$ 1.29 & $+0.05$ & 99.33 $\pm$ 0.21 & $-0.06$ \\
\midrule
\multicolumn{10}{l}{\emph{Qwen3 family}}\\
\quad Qwen3-0.6B & CLIP ViT-L/336 & \phantom{0}6.3 $\pm$ 2.1 & $+2.6$ & \phantom{0}2.7 $\pm$ 0.6 & $+1.7$ & 99.64 $\pm$ 0.08 & $-0.23$ & 99.10 $\pm$ 0.30 & $-0.37$ \\
\quad Qwen3-0.6B & MobileCLIP-S2  & \phantom{0}2.7 $\pm$ 2.1 & $-0.3$ & 14.7 $\pm$ 2.1 & $-0.6$ & 98.05 $\pm$ 0.28 & $+0.08$ & 99.61 $\pm$ 0.30 & $+0.04$ \\
\quad Qwen3-4B   & CLIP ViT-L/336 & \phantom{0}9.0 $\pm$ 8.7 & $+4.7$ & \phantom{0}1.3 $\pm$ 0.6 & $\phantom{+}0.0$ & 99.83 $\pm$ 0.08 & $\phantom{+}0.00$ & 98.71 $\pm$ 1.25 & $-0.67$ \\
\midrule
\multicolumn{10}{l}{\emph{Llama-3 family}}\\
\quad Llama-3.2-1B & CLIP ViT-L/336 & \phantom{0}4.7 $\pm$ 0.6 & $+4.0$ & \phantom{0}3.0 $\pm$ 1.0 & $+0.3$ & 99.60 $\pm$ 0.13 & $-0.04$ & 99.33 $\pm$ 0.08 & $-0.57$ \\
\quad Llama-3.2-3B & CLIP ViT-L/336 & 16.0 $\pm$ 2.6 & $+14.7$ & \phantom{0}0.3 $\pm$ 0.6 & $-1.7$ & 99.96 $\pm$ 0.08 & $+0.23$ & 97.71 $\pm$ 0.37 & $-2.11$ \\
\midrule
\multicolumn{10}{l}{\emph{Gemma-2 family}}\\
\quad Gemma-2-2B & CLIP ViT-L/336 & 10.3 $\pm$ 2.5 & $+5.3$ & \phantom{0}0.7 $\pm$ 0.6 & $-0.6$ & 99.91 $\pm$ 0.08 & $+0.08$ & 98.52 $\pm$ 0.36 & $-0.76$ \\
\bottomrule
\end{tabular}
}
\caption{Safety-relevant detection errors of the exported student models on the Detectium sensor. \(\Delta\) denotes the change relative to the corresponding student before export, as reported in Table~\ref{tab:confusion-ablation}.
 }
\label{tab:cpu-confusion}
\vspace{-2ex}
\end{table*}

\subsection{Export Pipeline}

\paragraph{Students with CLIP Encoder.}
Because the student models are trained using LoRA~\cite{hu2021lora}, each trained checkpoint consists of a PEFT LoRA adapter together with a two-layer MLP vision-language projector. We export each CLIP-based student using the \texttt{llama.cpp} toolchain~\footnote{\url{https://github.com/ggml-org/llama.cpp}} in four steps. First, the LoRA adapter is merged into the corresponding base language model to obtain dense float16 weights. Second, the merged language model is converted to GGUF format and quantized to Q8\_0. Third, the CLIP vision encoder and the student-specific two-layer MLP vision-language projector are packaged into a separate F16 multimodal projector file. Because the projector is trained independently for each student, this file cannot be shared across models. Fourth, the chat template used during training is embedded into the GGUF file so that prompts are formatted consistently with the training procedure at inference time. The template used for each model family is summarized in Table~\ref{tab:prompt-template}. We found this step to be important in practice: a mismatched chat template does not necessarily cause an explicit runtime error and can still produce fluent outputs, but substantially increases the false-alarm rate. For Qwen3 and Llama-3.2, the system message is:
\begin{quote}
\small
\texttt{You are a helpful language and vision assistant. You are able to understand the visual content that the user provides, and assist the user with a variety of tasks using natural language.}
\end{quote}

The final deployed configuration therefore consists of a Q8\_0-quantized language model together with an F16 multimodal vision module. At inference time, the exported files are served using \texttt{llama-server} from \texttt{llama.cpp}, which loads the model and performs inference directly on the sensor CPU.

\paragraph{Students with MobileCLIP Encoder.}
The \texttt{llama.cpp} conversion pipeline supports the CLIP-based vision encoder used in our main configurations, but does not support the FastViT~\cite{vasu2023fastvit} backbone used by MobileCLIP-S2. As a result, the MobileCLIP-S2 students, Qwen3-0.6B and Qwen2.5-3B, cannot be exported through the same GGUF pipeline. Instead, these models are executed directly on the sensor CPU using PyTorch in bfloat16 precision, with the LoRA adapter applied at load time and the vision encoder loaded through \texttt{open\_clip\_torch}~\cite{ilharco2021openclip,cherti2023reproducible}. The same four CPU threads are used as in the GGUF experiments.

Because the MobileCLIP-S2 models use a different runtime and numerical precision from the quantized GGUF models, their latency and accuracy results are not strictly directly comparable to those of the CLIP-based students. We therefore report them primarily to characterize the deployment behavior of the lightweight vision encoder under the available runtime.

\subsection{Measurement}

For evaluation, we use the same DetectiumFire evaluation set as in the main experiments. Each inference request contains a single previously unseen image together with the prompt corresponding to the fire-understanding task. We use deterministic decoding with temperature set to 0 and a maximum generation length of 64 new tokens. No image or prompt cache is reused across requests, so each request independently re-encodes the input image, matching the behavior of a sensor processing a live image stream.

We report three deployment metrics. \emph{Time to first token (TTFT)} measures the interval from receiving an inference request to generating the first output token. For a vision-language model, TTFT primarily captures the prefill stage, including image encoding, projection of visual features into the language-model embedding space, and the initial language-model forward pass. \emph{End-to-end latency (E2E)} measures the total inference time, including both TTFT and autoregressive generation of the remaining output tokens. \emph{Peak resident set size (RSS)} measures the maximum physical memory occupied by the inference process during execution. For each metric, we report the median over 20 randomly sampled fire-understanding requests from the validation set rather than the mean, following standard benchmarking practice for latency measurements~\cite{hoefler2015scientific}.

Qwen2.5-3B with MobileCLIP-S2 and Qwen3-4B require approximately 8.8 GB and 9.9 GB of memory, respectively, exceeding the memory capacity available on the target Detectium sensor. We therefore report their measurements for reference only. Neither configuration is deployable under the target hardware constraints.

\section{Additional Deployment Results}
\label{app:deploy-result}

Table~\ref{tab:cpu-confusion} decomposes on-device fire/non-fire detection error into false alarms and missed fires, together with recall and specificity. The results reveal more clearly how quantization changes the prediction behavior of the distilled students. For every CLIP-based student, the number of false alarms increases after export and specificity consequently decreases. Thus, \emph{the effect of quantization is not simply a general loss of accuracy, but a directional shift toward predicting fire more frequently}. In contrast, the two MobileCLIP-based students, which are executed without quantization, exhibit no comparable shift.

Moreover, \emph{the magnitude of this shift increases with model size}. Among the CLIP-based students, the four models with 1.5B parameters or fewer produce, on average, 2.7 additional false alarms after export, whereas the four models with 2B parameters or more produce 8.8 additional false alarms. Llama-3.2-3B provides the clearest example, producing 14.7 additional false alarms and losing 2.11 percentage points of specificity after export, compared with increases of 4.0 false alarms and a 0.57-point specificity drop for Llama-3.2-1B. These results suggest that larger students are more sensitive to quantization. Consequently, \emph{smaller students are not only faster and more memory-efficient, but also degrade less under the compression required for on-device deployment}.

The MobileCLIP-based students exhibit a different failure mode. Although they produce only 4.7 and 2.7 false alarms, respectively, among the lowest values in Table~\ref{tab:cpu-confusion}, they miss 39.3 and 14.7 fires. \emph{This failure mode is particularly concerning for real-world deployment}. False alarms produced by the CLIP-based models can still be resolved through human verification, whereas a missed fire may remain entirely undetected and therefore carries substantially greater safety consequences. Thus, despite their low false-alarm rates, the high number of missed fires makes the MobileCLIP-based configurations less suitable for practical fire detection.

\end{document}